\documentclass[letterpaper, 10 pt, conference]{ieeeconf} 
\IEEEoverridecommandlockouts 
\usepackage{tabularx}
\usepackage[latin1]{inputenc} 
\usepackage[cmex10]{amsmath}
\usepackage{algpseudocode}
\usepackage{amsfonts} 
\usepackage{amssymb} 
\usepackage{amsbsy}
\usepackage{hyperref} 
\usepackage{comment}
\usepackage[scriptsize,tight]{subfigure} 
\usepackage[]{units}
\usepackage{bigstrut}
\usepackage{times} 
\usepackage[pdftex]{graphicx}
\graphicspath{{./figures/}} \DeclareGraphicsExtensions{.pdf}
\usepackage[disable]{todonotes}

\newcommand{\Anote}[1]{\todo[inline,color=red!40]{\textbf{Alex:} #1}}
\newcommand{\VEC}[1]{\mathbf{#1}}
\newcommand{\symVEC}[1]{\boldsymbol{#1}}
\newcommand{\dVEC}[1]{\dot{\mathbf{#1}}}
\newcommand{\ddVEC}[1]{\ddot{\mathbf{#1}}}
\newcommand{\dsymVEC}[1]{\dot{\boldsymbol{#1}}}

\newcommand{\eqsref}[2]{(\ref{#1}-\ref{#2})}
\title{\LARGE \bf
 Trajectory generation for multi-contact
  momentum-control
 }
\author{Alexander Herzog$^{1}$, Nicholas Rotella$^{2}$, Stefan Schaal$^{1,2}$, Ludovic
  Righetti$^{1}$
  \thanks{\scriptsize This research was mainly supported by the
    Max-Planck-Society and the European Research Council (ERC) under
    the European Union's Horizon 2020 research and innovation
    programme (grant agreement No 637935). It was also supported by
    National Science Foundation grants
    ECS-0326095, IIS-0535282, IIS-1017134, CNS-0619937, IIS-0917318,
    CBET-0922784, EECS-0926052, CNS-0960061, the DARPA program on
    Autonomous Robotic Manipulation, the Army Research Office, the
    Okawa Foundation and the ATR Computational Neuroscience
    Laboratories.}
  \thanks{\scriptsize$^{1}$Autonomous Motion Department, Max Planck
    Institute for Intelligent Systems, %
    T\"ubingen, Germany. {\tt\small first.lastname@tuebingen.mpg.de}}
  \thanks{\scriptsize$^{2}$CLMC Lab, University of Southern
    California, Los Angeles, USA.} }

\begin{document}
\maketitle
\thispagestyle{empty}
\pagestyle{empty}

\begin{abstract}
Simplified models of the dynamics such as the linear inverted pendulum model (LIPM) have 
proven to perform well for
 biped walking on flat ground. However, for more complex tasks the
  assumptions of these models can become limiting. For example, the LIPM 
  does not allow for the control of contact forces independently, is limited to co-planar contacts
  and assumes that the angular momentum is zero.
  In this paper, we propose to use the full momentum equations of a humanoid robot in a trajectory optimization framework
  to plan its center of mass, linear and angular momentum trajectories. The model also allows
  for planning desired contact forces for each end-effector in arbitrary contact locations.
  We extend our previous results on LQR design for momentum control by computing the
  (linearized) optimal momentum feedback law in a receding
  horizon fashion.
  The resulting desired momentum and the associated feedback law are
  then used in a hierarchical whole body control approach. Simulation experiments
  show that the approach is computationally fast and is able to generate plans for locomotion
  on complex terrains
  while demonstrating good tracking performance for the full humanoid control.
\end{abstract}

\section{INTRODUCTION}

Humanoid robots locomoting and performing manipulation tasks on uneven ground
are required to actively apply forces on objects and
the floor in order to achieve their task successfully. A direct
effect of contact forces is a change of momentum in the robot, which
on the one hand is necessary to move the center of mass, but on the
other hand restricts the type of limb motion that the robot can
perform. The nonlinear nature of the angular momentum dynamics makes
preview-based control computationally hard in general and even with an
admissible angular momentum trajectory open control parameters
like the joint motion and feedback control remain problematic for use in a whole body control framework.

Successful applications of simplified momentum models have been shown
on robots walking quite robustly over flat ground\Anote{citations go
  here. Toro, DRC robots}. A common approach is to use the linear
inverted pendulum model (LIPM), which has been exploited for preview
control since it was introduced by Kajita et. al~\cite{Kajita:2003uh}. In~\cite{Herdt:2010bh}, a model predictive control (MPC) approach is formulated that finds foot steps on a flat ground together with a compatible CoM location on a horizontal plane. CoM profiles can then be realized on the full body together with other limb motion (e.g. swing leg) controllers~\cite{Sherikov:2014vq},~\cite{Feng:2013}~\cite{Faraji:2014tl},~\cite{herzog:2014b}.
The assumption of a horizontal CoM motion and flat ground can be relaxed to a pre-designed CoM height profile~\cite{Englsberger:2015jp},~\cite{Audren:2014gl}. Although, less restrictive than the original LIPM, these approaches are either built for point feet or leave parts of the dynamics uncontrolled. Depending on the terrain, pre-defining the CoM along a fixed direction may result in suboptimal or infeasible reaction forces.
Approaches that leave the regime of linear models have been proposed, for instance for long jumps~\cite{Wensing:2014bo} leading to more complex task behavior. However, they often require task specific models that for example take into account the swing leg dynamics. Going even further, we have seen work that optimizes over the whole joint trajectories and the full momentum together~\cite{Dai:2014tp}. The impressive near-to physical motions, however, come at a high computational cost.
Time-local controllers, have shown that
balancing performance of robots is improved, when overall angular
momentum is damped out directly~\cite{Lee:2012hb,Herzog:2014uv}. Nevertheless, it has remained unclear how angular momentum profiles should be chosen in non-static configurations.

In this paper, we consider the full momentum dynamics of the robot for generation of CoM and momentum trajectories and the according reaction force profiles at each contact. We formulate the problem as a continuous-time optimal control problem in a sequential form. A mode schedule is predefined together with end effector trajectories. A desired angular momentum trajectory, which is required for the limb motion, is generated from a simple inverse kinematics forward integration and realized with admissible contact forces in a least squares optimal sense. Since the kinematics information is used before the optimization over the dynamics, our optimization
procedure is relatively fast (for example compared to \cite{Dai:2014tp}).
In our previous work \cite{Herzog:2014uv} we have proposed a way to compute control gains for the momentum task using a LQR design approach. We showed that it was able to significantly improve performance for balancing tasks on a real humanoid. However, this approach was only used for stabilization. In this paper, we extend the approach to receding horizon tracking control.
We use the planned trajectories to generate feedback gains using a LQR design where we linearize the non-linear momentum dynamics around the optimized trajectory. In a simulated stepping task, our humanoid traverses a terrain with changing height and angled stepping stones successfully. A whole-body controller computes joint torques that realize feedback loops on momentum as well as the swing leg motion and respect additional balancing and hardware constraints and generates physically realizable contact forces. A good tracking of overall momentum is achieved using our receding horizon LQR design.\\
This paper is organized as follows. In Sec~\ref{sec:momentum_trajectory} we formulate an optimal control problem to obtain momentum trajectories of the robot. Resulting trajectories are controlled with a LQR feedback design as discussed in Sec~\ref{sec:momentum_lqr} which is then incorporated into a whole-body control approach as explained in Sec~\ref{sec:whole_body}. In Sec~\ref{sec:results} we demonstrate our control framework on a simulated stepping task on rough terrain. We discuss our results in Sec~\ref{sec:discussion} and finish with a conclusion in Sec~\ref{sec:conclusion}.

\Anote{
\begin{itemize}
\item Insist that the whole body controller was shown to work on a
  real robot and that is why we use it!a
\item \cite{Kudoh:2003gk} use inverted pendulum with momentum for his
  purposes, but the length of the pendulum remains fixed, which might
  be limiting in scenarios with several contact points.
\item \cite{Wensing:2014bo} ???
\item Once reaction force and momentum trajectories are computed, a
  local time controller \cite{Herzog:2014uv} can be used to track
  these. However, not any local time controller, but one that can
  control force tasks. Here inverse dynamics method are preferred.
\item With the momentum trajectory fixed, the planning problem of the
  whole body becomes a kinematic planning problem with only motion
  variables.
\item \cite{Kajita:2003gj},~\cite{Koolen:2012vh}~concepts for a
  stability criterion.
\item \cite{Grizzle:2014ub},~\cite{Posa:2014tg}~use the full dynamics
  for gait generation.
\item
  \cite{Sherikov:2014vq},~\cite{Herdt:2010bh},~\cite{Faraji:2014tl},~\cite{Kalakrishnan:2011dy}~use
  the linearized dynamics and optimize over a horizon.
\item \cite{Englsberger:2011jx}
\item \cite{Takao:2003fe}~suggest the use of the full momentum
  dynamics.
\item \cite{Ugurlu:2009fd} suggest to use the full momentum dynamics,
  but restrict feet to be flat on the ground.
\item \cite{Henze:2014ha} use the momentum dynamics in a MPC setting.
  However, only very short horizons are assumed and simplifying
  assumtions about the torque around the CoM are made.
\item Whole body controllers...
\item DRC...
\item Mordatch, Mansard, Escande, Mombaur, Diehl, Todorov, Leyvine
\item 2014 iros dlr stuff
\item \cite{Audren:2014gl}???
\item motivation:
  \begin{itemize}
  \item generalization of other approaches
  \item we can perform tasks that were limited by approximations
    before
  \item we are guided by experience on the robot, which is why we use
    hinvdyn and ctrl of momentum, since this is shown to work well on
    our robot.
  \item experiments in general show that for walking simple models are
    sufficient. Our goal is to find a simple model that is expressive
    enough to perform dynamic walking and simple enough to allow
    control design and high ctrl bandwidth.
  \end{itemize}
\end{itemize}
}

\section{MOMENTUM TRAJECTORY OPTIMIZATION} \label{sec:momentum_trajectory}

In this section, we describe how the robot momentum is planned together with admissible contact forces. We will first discuss the reduced dynamics model which will then be used to phrase an optimal control problem to generate CoM and momentum profiles. At the end of the section we describe how a desired angular momentum is chosen for optimization.

\subsection{Dynamic Model}
\begin{figure}
  \centering
  \includegraphics[width=\linewidth]{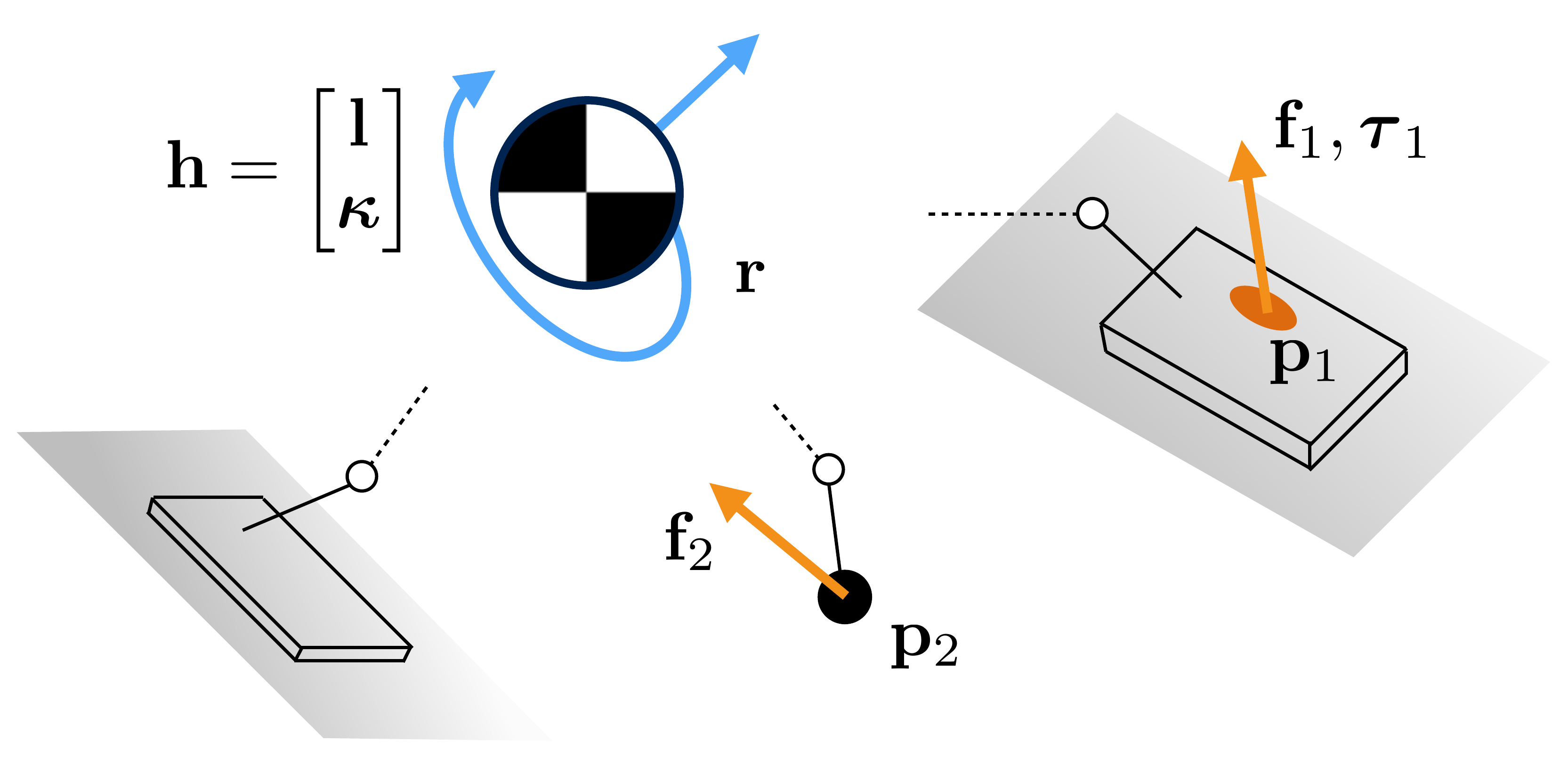}
  \caption{A sketch of the reduced model used through out this paper. It shows the center of mass
    (CoM) $\VEC{r}$ of the system and contacts with the
    environment. States and controls are color-labeled blue and orange respectively. Flat contacts that consist of a force, center of pressure and normal torque (here $\VEC{f}_1, \symVEC{\tau}_1, \VEC{p}_1$) can be modeled (e.g. a hand touching a wall) as well as point contacts like $\VEC{p}_2$. There is no restriction on the pose of contacts nor the CoM. Momenta generated around the CoM are not necessarily zero.}\label{fig:mass_and_contacts}
\end{figure}
We reduce the state of a robot to its center of mass
(CoM) $\VEC{r}$ and overall momentum
\begin{equation}
  \VEC{h} = \begin{bmatrix} \VEC{l} \\ \symVEC{\kappa} \end{bmatrix},
\end{equation}
with linear $\VEC{l}$ and angular $\symVEC{\kappa}$ momenta. The state
of the system is controlled by forces $\VEC{f}_i$ and torques
$\symVEC{\tau}_i$ acting at points $\VEC{p}_i$ on the mechanical structure. The momentum dynamics
as illustrated in Fig~\ref{fig:mass_and_contacts} can be written
\begin{eqnarray} 
\label{eq:momentum_dynamics_1}
  M\dVEC{r} &=& \VEC{l} \\
\label{eq:momentum_dynamics_2}
  \dVEC{l} &=& M\VEC{g} + \sum \VEC{f}_i \\
\label{eq:momentum_dynamics_3}
  \dsymVEC{\kappa} &=& \sum \symVEC{\tau}_i+ \sum (\VEC{p}_i -
                       \VEC{r})\times \VEC{f}_i,
\end{eqnarray}
where $M$ is the mass of the robot. The contact points $\VEC{p}_i$ can
for instance be point foot locations that are touching the ground or
they can represent the position of a handle that the robot holds on
to. Flat feet can be modeled by several contact points on the foot
surface. Equivalently, we can represent the effective force at the
center of pressure (CoP). Depending on the type of contact, it will be
necessary to express constraints on the forces. In the case of point
feet, no torques $\symVEC{\tau}_i$ can be generated. Torques that are
generated at the CoP of a flat foot are required to be normal to the
foot surface. Further, if we assume stationary contact points, we need
to restrict the wrench to remain in a friction cone.

\Anote{
The following is still missing:
\begin{itemize}
\item reaction force at CoP vs contact points
\item We can compute the unconstrained optimum
\item it is a generalization of the LIPM
\item relation to flywheel ???
\end{itemize}
}

\subsection{Optimal control formulation} \label{sec:trajectory_optimization}
\Anote{Maybe, first write very generally the quantities as function of
  time and also the constraints as function of time and objective
  function as integral over squared sum. From there introduce
  simplifications. Later show that further simplifications lead to
  existing approaches (piecewise linear functions, etc.)}
In the following we use the momentum dynamics of the robot to plan
for admissible contact forces that generate desired (linear and angular)
momentum trajectories. Our goal
is to compute force and contact point trajectories
$\VEC{f}_i(t), \symVEC{\tau}_i(t), \VEC{p}_i(t)$ that satisfy the
momentum dynamics in Eqs~\eqsref{eq:momentum_dynamics_1}{eq:momentum_dynamics_3} and contact constraints at all time. These trajectories are also required to steer the momentum through desired states over a time horizon $T$. 
Given the initial state of the robot CoM and momentum
$\VEC{r}(0), \VEC{l}(0), \symVEC{\kappa}(0)$, we can integrate
Eqs~\eqsref{eq:momentum_dynamics_1}{eq:momentum_dynamics_3} to obtain
\begin{eqnarray}
  \VEC{l}(t) &=& \int_0^t(\sum \VEC{f}_i(\delta) + M\VEC{g})d\delta,\label{eq:state_quantities_1}\\ 
  \VEC{r}(t) &=& \VEC{r}_0 + \frac{1}{M}\int_0^t \VEC{l}(\delta)d\delta,\\
  \symVEC{\kappa}(t) &=& \int_0^t (\sum \symVEC{\tau}_i(\delta) + \\ \nonumber
  &&\sum (\VEC{p}_i(\delta) - \VEC{r}(\delta))\times\VEC{f}_i(\delta))d\delta,\label{eq:state_quantities_5}
\end{eqnarray}
Note that the states are expressed as (nested) functions of the reaction forces. They are constructed by integrating, summing and applying the cross-product on $\VEC{f}_i(t), \symVEC{\tau}_i(t), \VEC{p}_i(t)$. Given a naive idea of what the desired CoM $\VEC{r}_{des}(t)$ and momentum $\VEC{h}_{des}(t)$ should be (e.g. coming from an initial kinematic plan), we want to find contact forces that minimize the error

\begin{eqnarray}\label{eq:cost}
J = &\sum_{t_0}^T(&
     \| \dVEC{l}(t_i)\|_{W_1}^2 + 
      \| \VEC{l}(t_i) - \VEC{l}_{des}(t_i)\|_{W_2}^2 + \nonumber \\ 
&& \| \VEC{r}(t_i) - \VEC{r}_{des}(t_i)\|_{W_3}^2 + 
      \| \dsymVEC{\kappa}(t_i)\|_{W_4}^2 + \nonumber \\ 
&& \| \symVEC{\kappa}(t_i) - \symVEC{\kappa}_{des}(t_i)\|_{W_5}^2~~~),
\end{eqnarray}

where we compute the errors at $t_i \in [0; T]$ and $W_i$ are diagonal weighting matrices that allow trade-offs between the
cost terms.

\subsubsection*{Adaptive end effector location}
Throughout the discussion in this paper we assume that the end effector moves between a series of predefined locations. However, this assumption can be relaxed without
making the optimization problem harder.
We simply substitute $\VEC{p}_i=\bar{\VEC{p}}_i +
\tilde{\VEC{p}}_i$, where $\bar{\VEC{p}}_i$ is the foot sole location
that remains stationary throughout a contact phase
and $\tilde{\VEC{p}}_i$ is the (time-varying) center of pressure inside
of the foot sole. In this reformulation both, $\bar{\VEC{p}}_i$ and $\tilde{\VEC{p}}_i$, will be
optimization variables and are chosen automatically. The support planes still have to be decided before-hand, e.g. using a dedicated acyclic contact planer \cite{bretl04, tonneau15}.

\subsection{Optimization procedure}\label{seq:optimization_procedure}
We will now formulate the described problem into an optimization problem. First, the reaction forces are formulated as weighted basis functions, more specifically as polynomials of the form

\begin{eqnarray}
  \VEC{f}(t; \VEC{w}) = \alpha(t) \sum_{k=0}^{N-1} w_kt^k  = \Phi^T(t)\VEC{w}, \\
  \alpha(t) =  \begin{cases}1 & \text{if } \text{in contact at } t\\ 0 & \text{else}\end{cases},
\end{eqnarray}

with weights $w_i$ and basis functions $t^k$ summarized in
vectors $\Phi(t), \VEC{w}$. We define the mode-scheduling variable $\alpha(t)$ which specifies contact activation and deactivation and is set before-hand, for example it could be obtained by a higher level planner. \Anote{cite contact planners here} Polynomials have the advantage of generating smooth force trajectories by construction.
The forces can then be written as

\begin{eqnarray}\label{eq:control_representation_1}
  f_i^j(t; \VEC{w}_i^j)           &=& \Phi^T(t) \VEC{w}_i^j,~j \in {x,y,z} \\ \label{eq:control_representation_2}
  \tau_i(t; \VEC{v}_i) &=& \VEC{n}_i \Phi^T(t) \VEC{v}_i \\ \label{eq:control_representation_3}
  p_i^j(t; \VEC{u}_i^j)          &=& \Phi^T(t) \VEC{u}_i^j,~j \in {x,y}
\end{eqnarray}

where the subscript identifies the end effector, the superscript
represents the coordinates and $\VEC{n}$ is the normal vector of the
foot sole. Contact forces, torques and CoPs thus become linear functions of polynomial coefficients. Expressing the states in Eqs~\eqsref{eq:state_quantities_1}{eq:state_quantities_5} with Eqs~\eqsref{eq:control_representation_1}{eq:control_representation_3} substituted, gives us the states as functions of time and polynomial coefficients. At the core of operations required to carry out the result are multiplication, summation and integration of polynomials, which can be computed analytically. As a result, we phrased our optimal control problem in sequential form, where our controls evaluate directly to states and the dynamics equations (cf. Eqs~\eqsref{eq:momentum_dynamics_1}{eq:momentum_dynamics_2}) are implicitly incorporated. This allows us to phrase our optimization problem

\begin{alignat}{5}\label{eq:optimization_problem}
  & \underset{\VEC{x}}{\text{min.}}~ 
  & 		&J(\VEC{x}) \\
  & \text{s.t.}  && | p^j(t_i;\VEC{x}) | \leq \hat{p},~j = x,y \label{eq:cop_constr}\\
  &                 && | \tau(t_i;\VEC{x}) | \leq \hat{\tau}, \label{eq:trq_constr}\\
  &                 && 0 \leq -| f^j(t_i;\VEC{x}) | + \mu f^z(t_i;\VEC{x}) \leq \hat{f_z}, ~j = x,y \label{eq:fric_constr} \\
  &                 &&t_i=t_0\dots T \nonumber
\end{alignat}

where $\VEC{x}$ is a concatenation of variables $\VEC{w}_i^j, \VEC{v}_i, \VEC{u}_i^j$. We try to find polynomial coefficients $\VEC{x}$ that minimize the error cost $J$ while satisfying friction and support bound constraints on forces. Eqs~\eqsref{eq:cop_constr}{eq:trq_constr} bound the CoPs and torques. In Eq~\eqref{eq:fric_constr} we impose a friction cone constraint approximated as pyramid and upper bound it by a sufficiently large value $\hat{f_z}$ to keep the polynomials from penetrating the lower bound and escaping above. In our constraints in Eqs~\eqsref{eq:cop_constr}{eq:fric_constr}, we wish to have bounds that hold at all $t \in [0, T]$; in practice however, we express them at a finite number of time steps. Given the limited flexibility of polynomials, we get minor penetration of those constraints in intervals $(t_i; t_{i+1})$.  Note that all our constraints are linear, whereas the objective function has quadratic terms as well as higher order (non-convex) terms due to the costs on angular momentum. Since the dynamics are not further simplified, we have the benefit that all states and controls are included in the cost $J$ and none of them are left uncontrolled.
\subsubsection*{Receding Horizon} Given that the problem in Eq~\eqref{eq:optimization_problem} is non-convex, we cannot expect to find a global optimum in general but need to find a local optimum starting from an initial guess $\VEC{x}_0$. In our approach we use a receding horizon technique. We start out by solving our problem for a horizon $\tilde{T} < T$ and obtain an optimal solution $\VEC{x}_{[0,\tilde{T}]}^*$. Typically, we start our desired motion such that it is easy to solve in the interval ${[0,\tilde{T}]}$, i.e. the robot is standing still and applying gravity compensation is already optimal. Then, we formulate our problem for the interval $[\Delta, \tilde{T}+\Delta]$ where we initialize our optimizer with the previous solution. This has the advantage that an initial solution can be bootstrapped from an initial easier-to-solve configuration and then moved over the horizon to the difficult parts of the motion. Further, as we push our implementation to run in real-time, we seek to use it in a receding horizon control setting.
\subsubsection*{Desired Angular Momentum} Most simplified momentum models assume that angular momentum is desired to remain at zero. This, however, ignores the fact that momentum might be required to move the robot limbs. E.g. in a stepping task we need to swing a leg, which in turn generates a momentum around the hip. Thus, it is not trivial to decide what the desired angular momentum should look like. In order to overcome this issue, we integrate forward our desired swing leg trajectories using inverse kinematics. From the resulting joint motion we then compute $\VEC{r}_{des}(t), \symVEC{\kappa}_{des}(t)$ (explained e.g. in~\cite{Orin:2008ge}). Although dynamically not feasible, this method generates angular momentum profiles required to perform the task-imposed motion, e.g. swinging a leg. We iterate between kinematic forward integration of an optimized $\VEC{r}(t)$ and optimizing a desired $\symVEC{\kappa}_{des}(t)$ with dynamic constraints (by solving the problem in Eqs~\eqsref{eq:optimization_problem}{eq:fric_constr}). In our experiments two passes already lead to convergence, i.e. forward integrated and optimized trajectories do not differ significantly. With this iteration we not only generate admissible force profiles that obey the momentum dynamics equations but also we can bootstrap angular momentum trajectories which are not obvious to design otherwise. Note that neither CoM nor momentum are predefined in advance. Instead an initial guess is given and the final trajectories are found automatically.

\section{MOMENTUM LQR} \label{sec:momentum_lqr}
In the previous section we described how reference trajectories in accordance with the
momentum dynamics are obtained. In order to track those trajectories on the full robot, we propose a feedback law using a LQR design, which has shown superior performance compared to a naive PD gain approach in experiments on the real robot~\cite{Herzog:2014uv}. From our trajectory optimizer we obtain admissible states and controls

\begin{eqnarray}
  \VEC{y}^* &=& \begin{bmatrix} \VEC{r} \\ \VEC{l} \\ \symVEC{\kappa} \end{bmatrix},
  \symVEC{\lambda}^* = \begin{bmatrix} \vdots \\ \VEC{f}_i \\ \symVEC{\tau}_i \\ \vdots \end{bmatrix},\\
  \dVEC{y}^* &=& f(\VEC{y}^*, \symVEC{\lambda}^*) = 
  \begin{bmatrix}
    \frac{1}{M}\VEC{l} \\
    M\VEC{g} + \sum \VEC{f}_i \\
    \sum \symVEC{\tau}_i+ \sum (\bar{\VEC{p}}_i - \VEC{r})\times \VEC{f}_i,
  \end{bmatrix} \label{eq:LQR_nonlin_dynamics}
\end{eqnarray}

where we transform wrenches $\VEC{f}_i,\symVEC{\tau}_i$ to the stationary poles $\bar{\VEC{p}}_i$.\\
The dynamics function in Eq~\eqref{eq:LQR_nonlin_dynamics} is discretized and linearized around the desired trajectories $\VEC{x}^*, \symVEC{\lambda}^*$. The resulting time varying-linear dynamics are then used to formalize a finite horizon LQR problem. This yields a control policy 
\begin{eqnarray}\label{eq:lqr_force_gain}
\symVEC{\lambda}=
\symVEC{\lambda}^* - \VEC{K}_t(\VEC{x} - \VEC{x}^*)
\end{eqnarray}
with time-varying feedforward and feedback terms that map errors in states into contact wrenches. Controlling the momentum with this feedback law requires 6 DoF for wrenches at each contact. Since our whole-body controller incorporates additional control objectives and force constraints, we compute directly the momentum rate that the wrenches generate

\begin{eqnarray}\label{eq:momentum_force_gain}
\dVEC{h}_{ref} = 
  \begin{bmatrix}
    \mathbf{I}_{3\times 3} & \mathbf{0}_{3 \times 3} &\ldots	\\
    [\bar{\VEC{p}}_i -
    \mathbf{r}^*]_{\times} &\mathbf{I}_{3\times 3}& \ldots
  \end{bmatrix}
                                                        \VEC{K}_t(\VEC{x}^*-\VEC{x}) + 
                                                        \dVEC{h}^*,
\end{eqnarray}

\Anote{check if it is K(x-x^*) or K(x^* - x)}

where $[\centerdot]_\times$ turns a cross-product into a matrix multiplication. The resolution of momentum rate to contact wrenches is then left to our whole-body controller as described in the next section. In this LQR design a quadratic performance cost is set once and then optimal gains are computed at each time step for the corresponding contact configuration. As we discussed in our previous work, in order to achieve compatible results with diagonal PD gain matrices, we had to design gains for different contact configurations, whereas the LQR design requires one performance cost and generated suitable gains automatically. In contrast to our previous work we linearize around desired trajectories, whereas in our robot experiments we used only two key configurations; this may become limiting in more versatile tasks such as walking over a rough terrain.
\Anote{Show how this is a better design compared to fixed gain
  matrices, especially PD control. E.g. tune a PD controller and
  evaluate the cost according to the LQR cost.}
\Anote{Right now we tune the QR matrices in an additional step. Should
we take them from the objective function? (We could approximate the
hessian with finite differencing and use eigenvalue decomposition to
obtain the psd part. We can then claim that the ev decomp can be
dropped, since the psd part of the hessian can be computed with auto diff.)}

\section{WHOLE-BODY CONTROL}\label{sec:whole_body}
The trajectories generated with the model in Eqs~\eqsref{eq:momentum_dynamics_1}{eq:momentum_dynamics_3} define the CoM, momentum and end effector forces of our humanoid. In order to track these trajectories on the full robot, we need to generate joint torques accordingly and at the same time control the limb motion and guarantee that other constraints are obeyed, e.g. joint limits. For this time-local control problem, we use inverse dynamics in QP Cascades, which we applied successfully on the real robot in previous work~\cite{herzog:2014b}. It allows us to phrase feedback controllers and constraints as functions of joint accelerations $\ddVEC{q}$, external generalized forces $\symVEC{\lambda}$ and joint torques $\symVEC{\tau}$. For instance, we can express a cartesian controller on the swing foot as an affine function of joint accelerations or a momentum controller as a function of reaction forces. The latter, for example, would be

\begin{eqnarray}
%
  \begin{bmatrix}
    \mathbf{I}_{3\times 3} & \mathbf{0}_{3 \times 3} &\ldots	\\
    \label{eq:mom_rate_frcs} [\bar{\VEC{p}}_i -
    \mathbf{r}]_{\times} &\mathbf{I}_{3\times 3}& \ldots
  \end{bmatrix}
                                                        \boldsymbol{\lambda} + 
                                                        \begin{bmatrix}
                                                          M\mathbf{g}
                                                          \\
                                                          \mathbf{0}
                                                        \end{bmatrix}
  = \dot{\mathbf{h}}_{ref}.
\end{eqnarray} 

A reference momentum rate $ \dot{\mathbf{h}}_{ref}$ can be chosen from a LQR design as demonstrated in our previous work ~\cite{Herzog:2014uv} and extended in Sec~\ref{sec:momentum_lqr}.
Given a set of controllers and tasks, we find torques that satisfy the dynamics equations of the full robot and at the same time generate the desired task feedback as good as possible. Tasks, however, may conflict, for instance moving the CoM may require moving the swing leg and vice versa. In these cases, QP cascade allow for two types of trade-offs; we can either weigh tasks against each other or we can prioritize them strictly. Especially when it comes to trade-offs between constraints, tasks of interest and redundancy resolution, prioritization can facilitate successful control design.\\

\section{SIMULATION RESULTS}\label{sec:results}
\begin{figure}
  \centering
  \includegraphics[width=\linewidth]{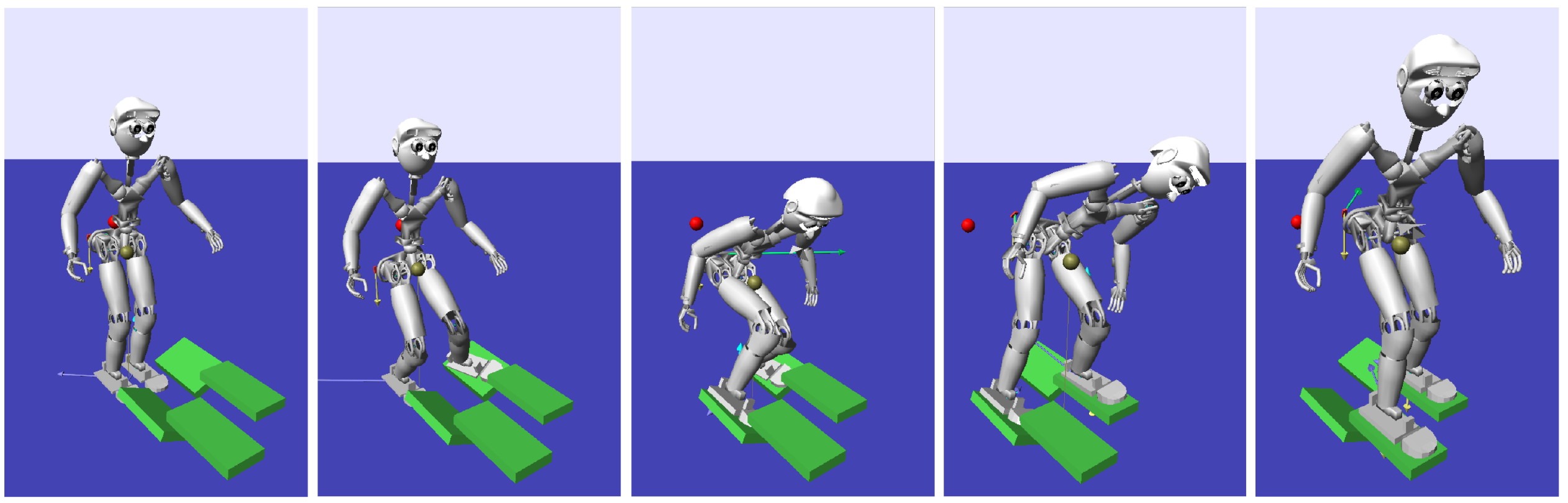}
  \caption{The humanoid robot traversing a terrain with stepping stones of different hight and orientation. }\label{fig:task_snapshots}
\end{figure}
\begin{figure}
  \centering
  \includegraphics[width=\linewidth]{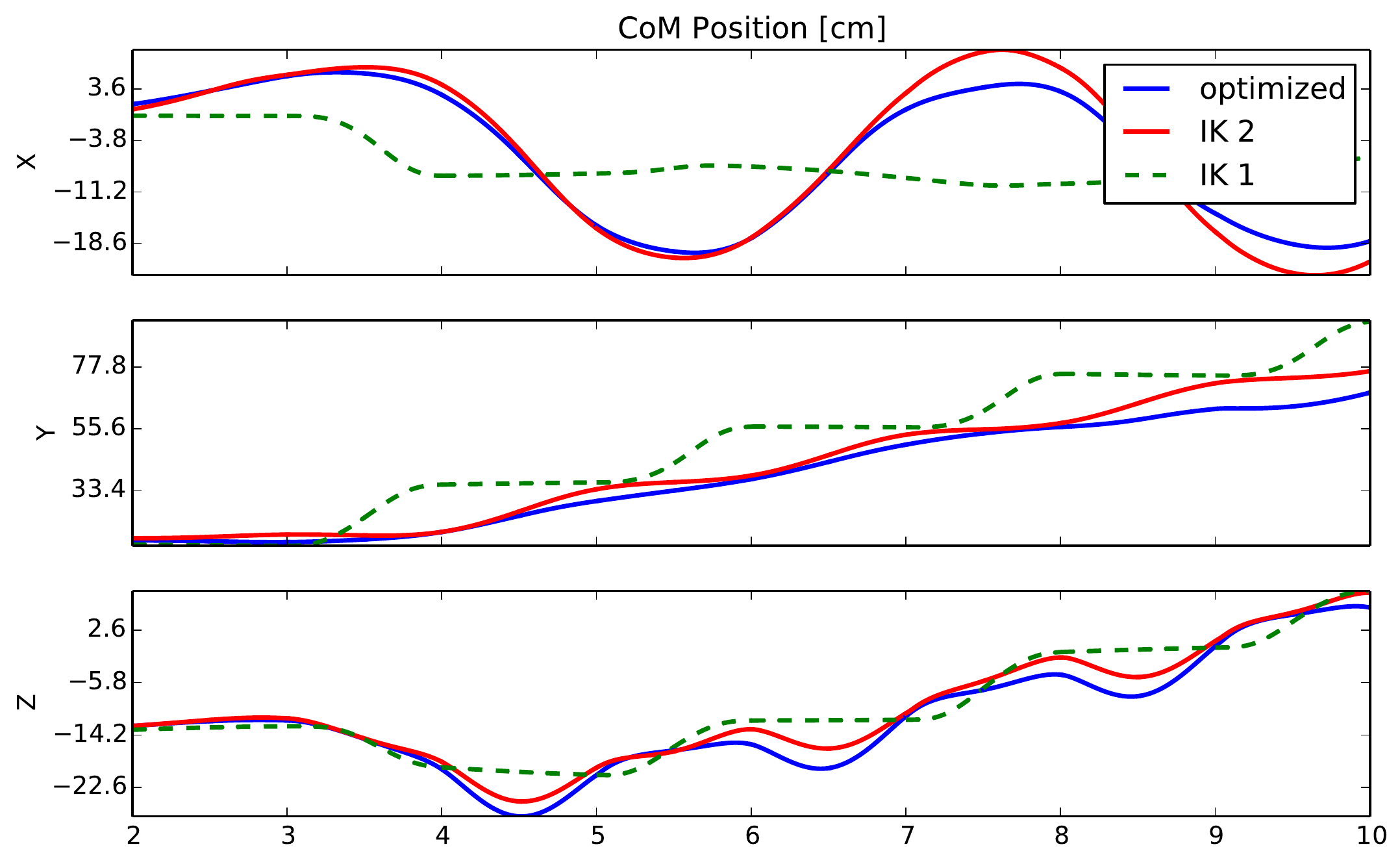}
  \includegraphics[width=\linewidth]{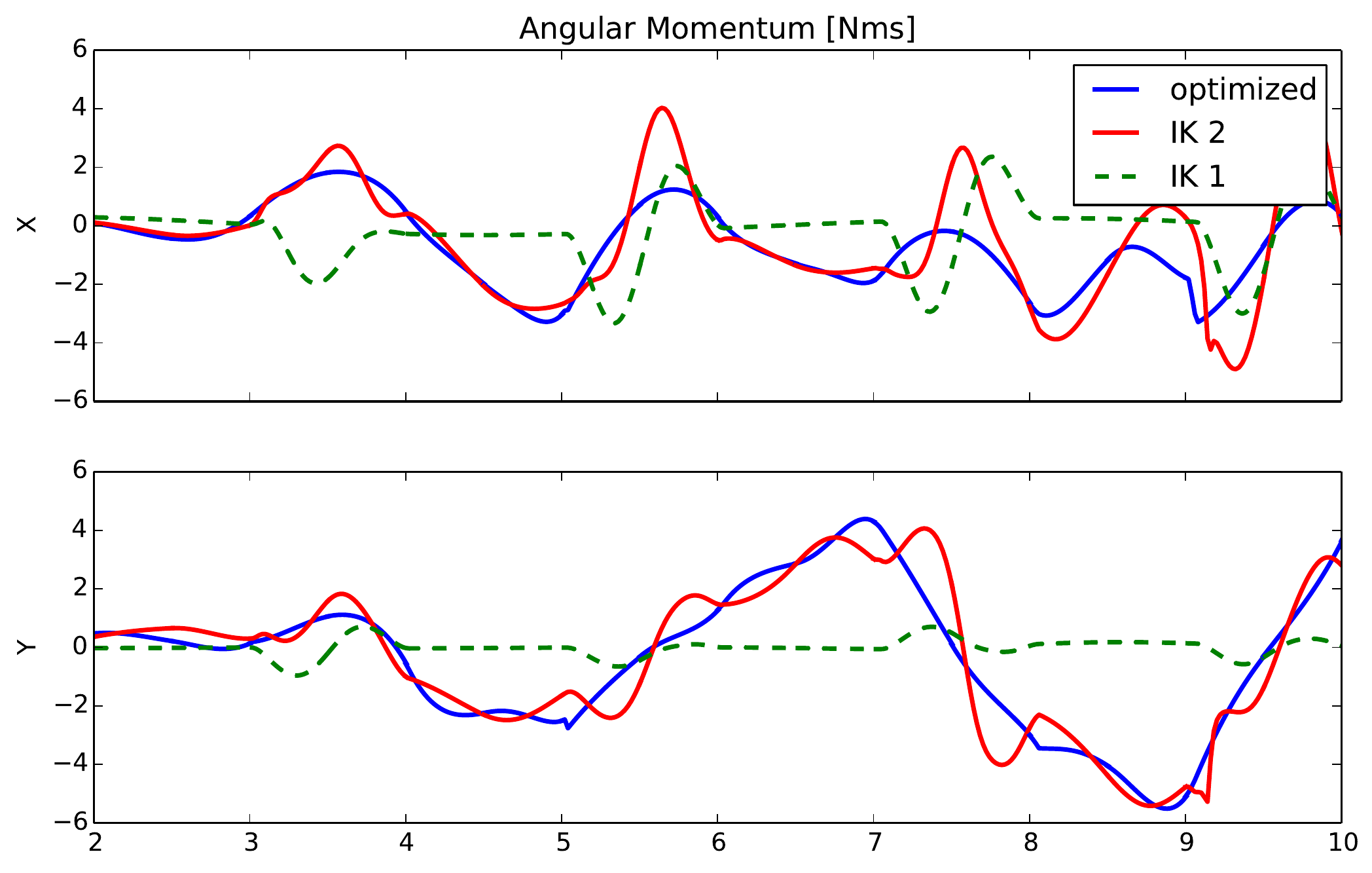}
  \caption{Plans for the CoM and horizontal angular momentum. {\it IK 1, IK 2} show the profiles obtained from the inverse kinematics passes, whereas the blue graph is the final optimized trajectory. The resulting lateral CoM (top plot) was adapted quite significantly over the planning process to make it physically compatible with the reaction forces. Angular momenta are found that allow for stepping motions required to traverse the terrain.}\label{fig:momentum_plan}
\end{figure}
\begin{figure}
  \centering
  \subfigure[Momentum LQR Gain, Eq~\eqref{eq:momentum_force_gain}]{
\begin{minipage}{0.5\linewidth}%
\vspace{1,68cm}
\includegraphics[width=\linewidth]{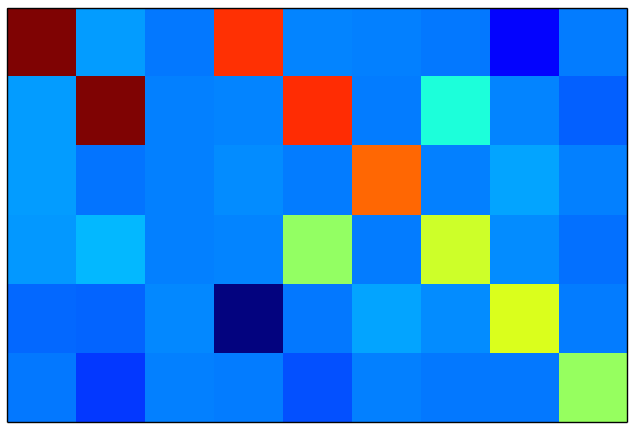}
\end{minipage}}%
\subfigure[Force LQR Gain, Eq~\eqref{eq:lqr_force_gain}]{\begin{minipage}{0.5\linewidth}%
\includegraphics[width=\linewidth]{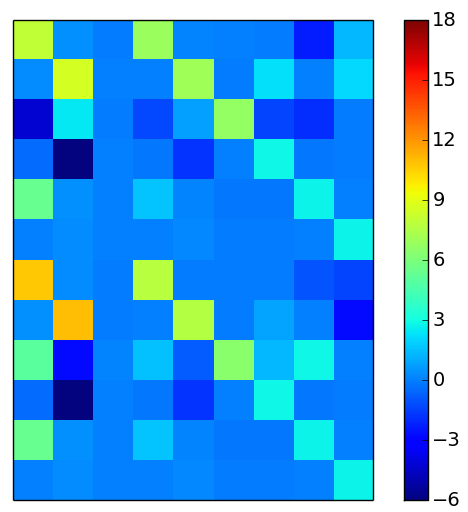}
\end{minipage}}
\includegraphics[width=\linewidth]{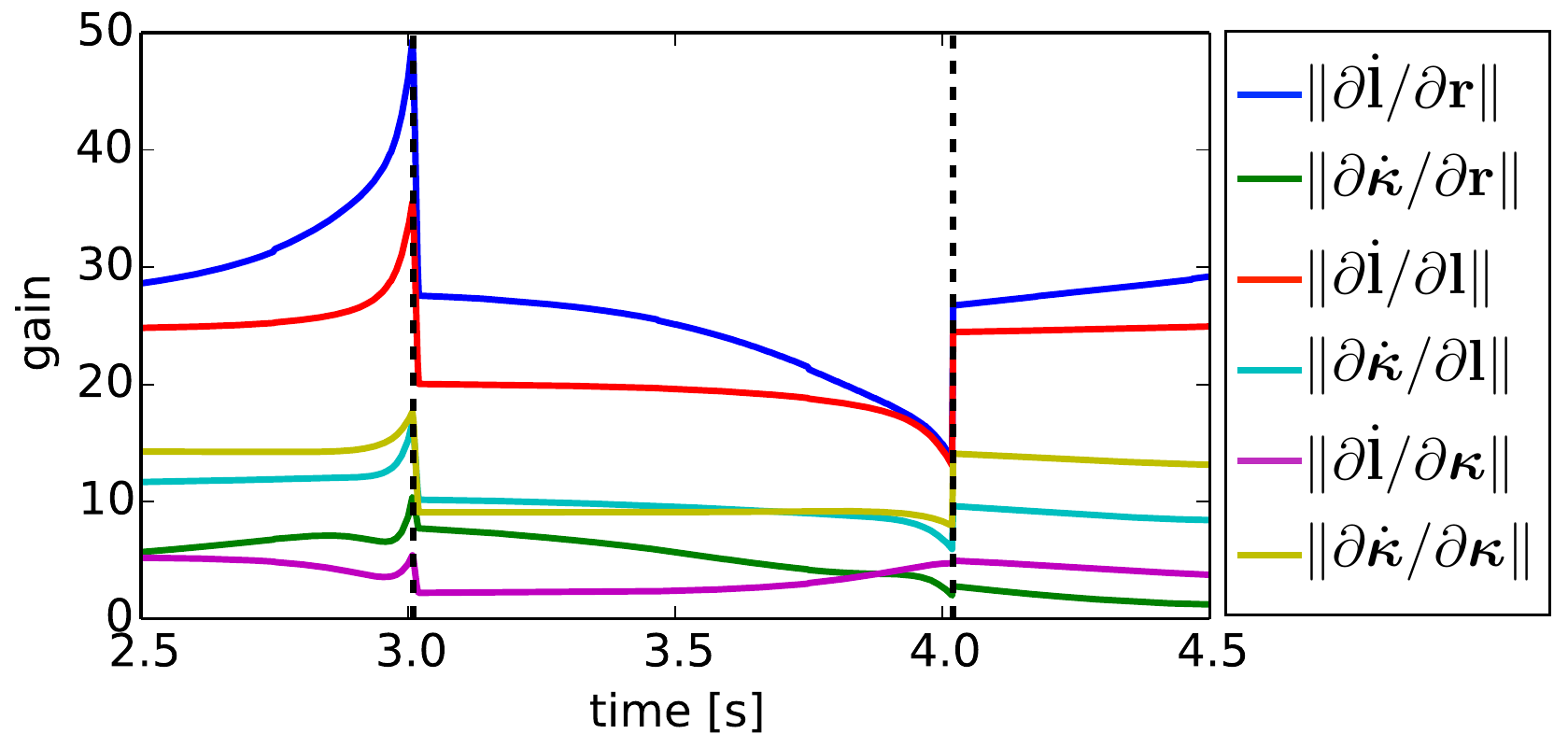}     
  \caption{Here we show a heat map visualization of {\it(a)} a momentum gain and {\it(b)} a force gain. The top 6 rows in {\it(b)} correspond to the wrench of the right foot, whereas the bottom rows correspond to the left foot. As we can see, the gains contain off-diagonal terms leading to coupling terms between linear and angular momentum. These terms are ignored in a naive diagonal PD gain design. The bottom plot shows the norm of 3x3 sub-blocks of the momentum gain plotted over time. The vertical dashed lines at $t=3$ and $t=4$ indicate contact switches. Gain profiles change significantly over time and contact configurations. The gains and momentum dynamics are discontinuous at contact switches. Nevertheless, discontinuities in joint torques were negligible. }\label{fig:momentum_gain}
\end{figure}
\begin{figure}
  \centering
  \includegraphics[width=\linewidth]{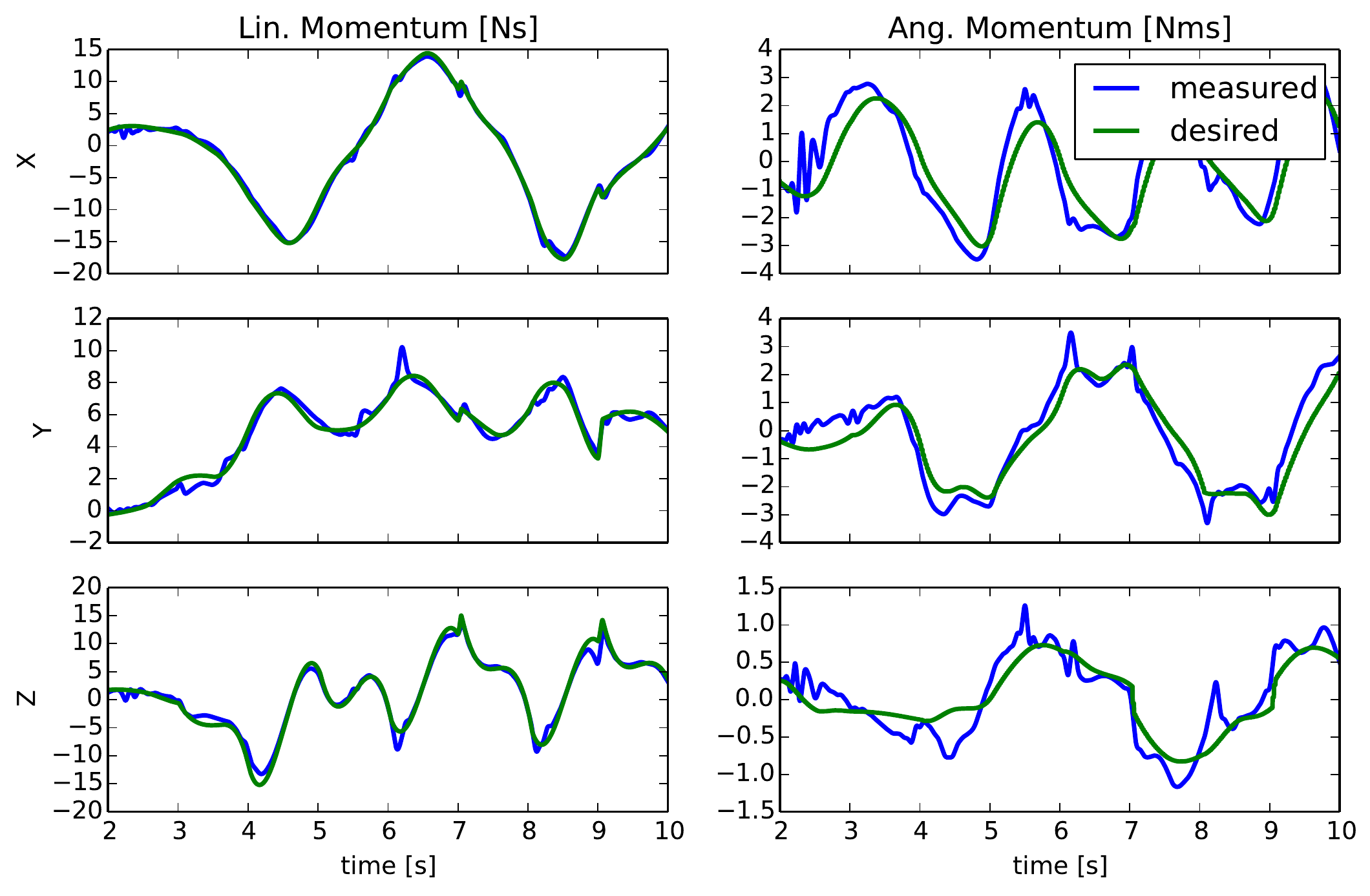}
  \caption{Linear and angular momentum are tracked well as the robot walks over the terrain. The oscillations in the angular momentum at t=2 come from rapid arm motions when the robot was trying to move the CoM up and the knees were stretched. This can be avoided, e.g. by adding box constraints on the CoM in the trajectory optimization step to account for kinematic limits.}\label{fig:mom_tracking}
\end{figure}
\begin{figure}
  \centering
  \includegraphics[width=\linewidth]{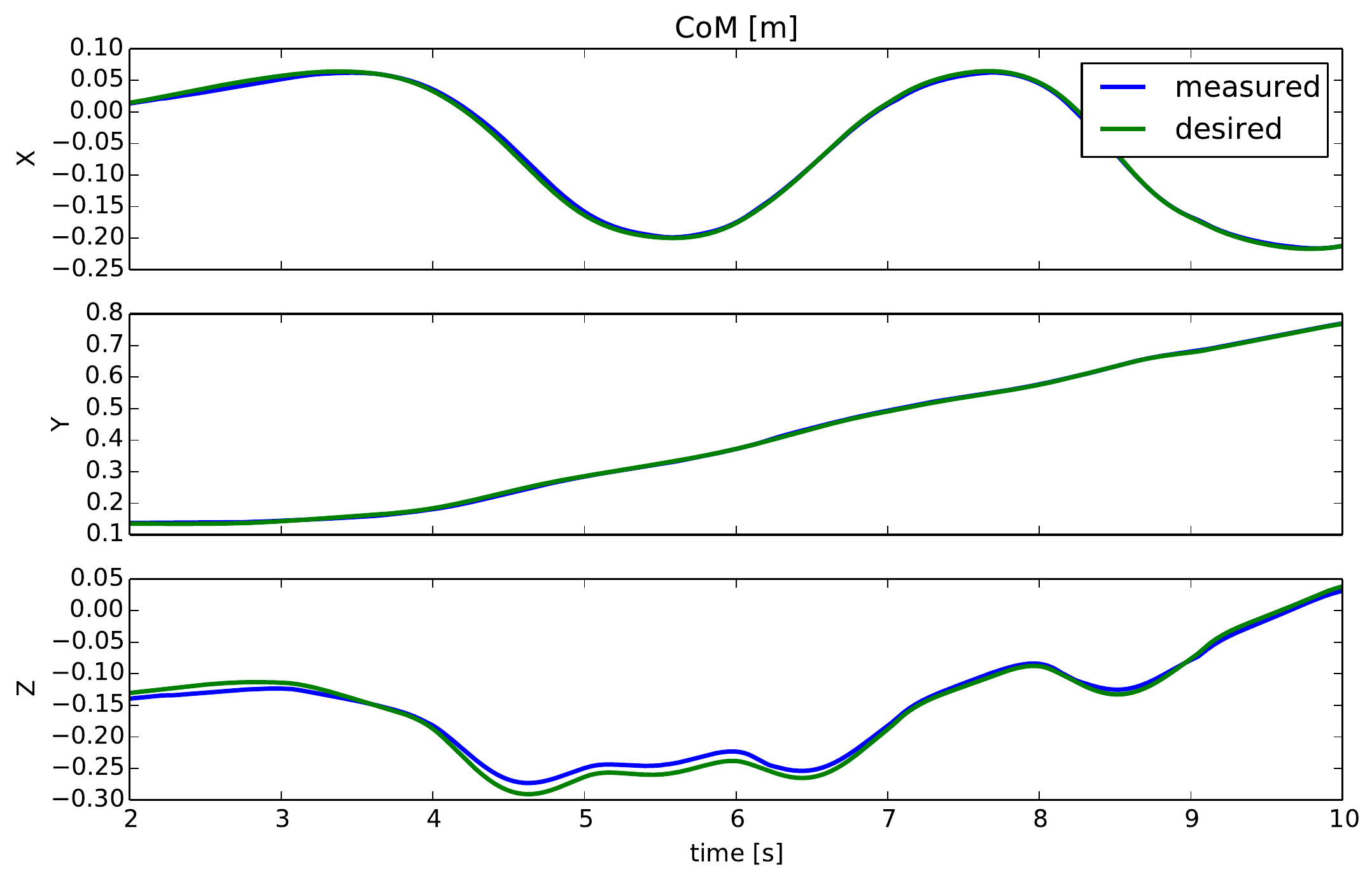}
  \caption{Measured and desired CoM of the robot when it was traversing the terrain. As the plots show, good tracking performance can be achieved with the proposed LQR design on the momentum.}\label{fig:com_tracking}
\end{figure}

This section describes our simulation results of the proposed control
framework. We use a model of our Sarcos humanoid in the SL simulation environment. Contacts are simulated with a penalty method and stiff springs. All experiments are performed on a 2.7 GHz intel i7 processor with 16gb ram. 
A task is generated where the robot is to walk on stepping stones that increase in height from one step to the other as visualized in Fig~\ref{fig:task_snapshots} and shown in the attached video\footnote{The video is also available on \url{www-amd.is.tuebingen.mpg.de/~ herzog/15_07-Humanoids.mp4}}. 
The z-axis of the inertial frame points up and the robot walks along the y-axis to the front. Two out of the four steps are tilted by $25^o$. Both the change in CoM height as well as the angled supports break the assumptions made in LIPM models and require the consideration of two separate contact wrenches. We generate swing leg trajectories using cubic splines to parameterize the pose of the foot. The humanoid starts out in double support at rest. After the first 3 seconds the left foot breaks contact and moves to an angled support surface located to the front left-hand side of the robot at an increased height (as shown in Fig~\ref{fig:task_snapshots}). Then a contact switch occurs at every second, changing from single support to double support or vice versa. The second step is again a support surface angled inwards to the robot and located to the front right-hand side. Finally, the robot takes two steps onto a horizontal plateau located slightly below knee height.\\
The planner is initialized with a naive idea of the robot motion where we simply keep the base at a certain height above the feet. After integrating the desired swing foot positions together with the base height using inverse kinematics (as described in Sec~\ref{seq:optimization_procedure}), we obtain resulting $\VEC{r}_{des}(t), \symVEC{\kappa}_{des}(t)$ trajectories that take into account the angular momentum required to swing the legs from one stepping stone to the other. The inverse kinematics solution, which is physically not consistent, is then adjusted in the trajectory optimization step to be admissible with respect to constrained forces and CoPs. None of the force constraints were notably violated. As can be seen in Fig~\ref{fig:momentum_plan} the inverse kinematics-generated CoM trajectory is modified significantly in the lateral direction and as a consequence the angular momentum in the y-direction is modified as well. After a second iteration of inverse kinematics integration and trajectory optimization, the results have sufficiently converged and we stop. Polynomials of order 3 are chosen and initialized with zero.  The planning process took 4 min and converged after only two iterations. This allows us to generate a complex motion rather quickly compared to motion planners that are based on more extensive models of the robot. The numerical optimization problem in Eq~\ref{eq:optimization_problem} is solved with SNOPT~\cite{Gill:2002}, a Sequential Quadratic Programming method.\\ 
Next, we construct a hierarchy of feedback controllers and constraints in order to realize the momentum profile on the full humanoid. At the highest priority we express the physical model of the full robot to obtain physically-consistent torques. This is followed by force and joint limits together with contact constraints. In the third priority we control the swing leg motion and the momentum and add a posture PD control with a relatively-low weight. In the priorities below, we regularize end effector forces and stabilize the base orientation. A summary of the task setup is given in Tab~\ref{tab:task_hierarchy}. %
\begin{table}
  \center
  \begin{tabularx}{\linewidth}{p{.1\linewidth}p{.27\linewidth}X}
    \hline
    \textbf{Rank} & \textbf{Nr. of eq/ineq constraints}  & \textbf{Constraint/Task} \\
    \hline
    1 &$6$ eq& Newton Euler Equation\\
    2 &$2 \times 6$ eq& Contact constraints\\
                  &$2\times 4$ ineq& Center of Presure\\
                  &$2\times 4$ ineq& Friction cone\\
                  &$2 \times 14$ ineq& joint acceleration limits\\
    3 &$6$ eq& LQR momentum control\\
                 &$6$ eq& Cartesian swing foot control\\
                 &$14+6$ eq& PD control on posture\\
    4             &$2 \times 6$ eq& Contact forces control\\
    5             &$3$ eq& Base link orientation\\
    \hline
  \end{tabularx}
  \caption{Hierarchy of tasks in the stepping experiment} 
  \label{tab:task_hierarchy}
\end{table}
The momentum is controlled with feedback gains designed as described in Sec~\ref{sec:momentum_lqr}, where our performance cost is fixed for the whole run. We penalize state errors by $10$, forces by $0.1$ and torques by $0.5$. Gains are generated over a 2sec horizon with a granularity of 200 time steps. A sequence of gains is recomputed every 10ms.\\
The robot was able to traverse the terrain successfully. The momentum trajectories, which were both dynamically consistent as well as compatible with the robots limb motions, could be tracked well as shown in Figs~(\ref{fig:com_tracking}-~\ref{fig:mom_tracking}). Torques were generated that are in the bounds of the physical robot's torque limits. In the beginning of the task the planned CoM height forces the knees to stretch, which prevents the robot from using its knees to lift the CoM further, but instead it lifts the arms up rapidly. This can be avoided by adding a box constraint on the CoM in the optimization problem in order to consider kinematic limitations.\\
The LQR gain matrices have non-zero off-diagonal values as expected (cf. Fig~\ref{fig:momentum_gain}). For instance, we can see that angular momentum is generated in order to correct for errors in CoM and linear Momentum, which would not be possible with diagonal PD gains. In fact, we tried to track the planned motion using diagonal PD gains. We started with values similar to the diagonal of the LQR gain, but we could not find parameters that were stable throughout several contact situations as was the case for the LQR gains. 
Increasing the terrain difficulty was mainly problematic due to kinematic limits because our naive swing leg trajectories required to keep the heel on the ground during the whole support phase, thus limiting the stepping height. Further, we noted problems with the simulator's contact model, which caused sporadic spikes in the force profile when we applied strong forces on the ground.\\
Overall, a complex task could be planned quickly including a non-trivial angular momentum profile that respected the end effector motion. The proposed feedback control law on the momentum showed good performance when it was embedded in an inverse dynamics task hierarchy.
 
\Anote{
- mention that initial IK CoM height and planned height differ.
- mention speed of planner.}

%

\section{DISCUSSION}\label{sec:discussion}
\Anote{
\cite{Posa:2013tg} are not constrained by mode scheduling, since it
can result in growth of complexity, when accounted for joint limit
hits, transitioning between static and sliding contacts, etc. However,
in our scenario we use a simplified model with only static contacts
and we do not take into account kinmaitcs (ant thus no jointlimits
exist). However, we can add (linear) inequality constraints in order
to put a maximum distance between center of mass and ankle.
}
\subsubsection*{Relation to simplified Momentum Dynamics}
The momentum model in
Eqs~\eqsref{eq:momentum_dynamics_1}{eq:momentum_dynamics_3} is
often simplified further in order to obtain linear dynamics, which then
leads to computationally more efficient algorithms. However, turning multiplicative terms between variables in
Eq~\eqref{eq:momentum_dynamics_3} into linear terms requires
potentially restrictive assumptions. E.g. in the LIPM the CoM height $r_z$ is assumed
to be constant, there is one effective $\VEC{p}_i$, which lies in one
horizontal plane, and the force is required to act along
$\VEC{r}-\VEC{p}$.\\
Since a constant CoM height may be limiting for tasks that require vertical movement, the authors in~\cite{Audren:2014gl} allow a predefined (not constant) $r_z(t)$.
Substituting this assumption into Eq~\eqref{eq:momentum_dynamics_3},
and assuming constant $\VEC{p}_i$, turns the horizontal angular
momentum into a linear function of forces and torques. Assuming in addition
that the nonlinear vertical angular momentum can be neglected, the
authors end up again with linear dynamics.
If we restrict our dynamics model further into a LIPM and use discretized dynamics (instead of polynomial trajectories) we can recover the approach of \cite{Herdt:2010bh}.
However, this can be restrictive if we want to step on various slopes
or when a non-zero angular momentum is required, e.g. to consider limb motion. 

\subsubsection*{Computation time}
In our current work, we focused on the capabilities of the
multi-contact dynamics model to generate dynamic motions on uneven
terrain and we showed that these behaviors can be controlled on a full
humanoid robot. However, the goal of separating the control process
into a predictive control generation on a lower dimensional model and time-local control on the full dynamics has the potential for a fast implementation in a MPC fashion. In our experiments we saw potential drawbacks in our numerical optimization in Eq~\eqref{eq:optimization_problem}. Increasing the order of polynomials leads to slower convergence rates, whereas reducing the polynomial dimensions too far may lead to poor flexibility of the trajectory representation. This may be explained by the discrepancy of basis functions evaluated close to 0 and close to T leading to poorly conditioned problem. Piecewise constant trajectories may overcome this problem. Another point for improvement of the numerical procedure is the formulation of the optimal control problem. In the current formulation, the objective function in Eq~\eqref{eq:cost} has terms up to 4th order (squared norm of cross products). Our numerical solver is based on approximations up to second order, which may limit the region in which approximations are valid. It is possible (but out of the scope of this paper) to rewrite the problem into a (non-convex) Quadratically Constrained Quadratic Program where the objective function as well as the constraints are quadratic, leading to a better approximation in second order methods and potentially improving convergence. Pushing the implementation towards an online control algorithm is part of our future work. 

\section{CONCLUSIONS}\label{sec:conclusion}
We presented an approach to control contact forces and momentum for humanoid robots. CoM and momentum profiles were obtained in an optimal control framework together with admissible contact forces. Feedback gains are generated from a LQR design, which generates time and contact-configuration dependent gains from a single performance cost. The resulting controller is embedded in an inverse dynamics-based whole-body controller together with other limb controllers and constraints. We demonstrated the control framework on a simulation of the Sarcos humanoid traversing rough terrain. Physically admissible momentum and force trajectories could be found relatively quickly and were tracked well during the task execution.
In future work, we will implement the discussed speed-up and push the optimal controller to run online.


\bibliographystyle{IEEEtran} {\footnotesize
  \bibliography{humanoids2014} }

\begin{thebibliography}{10}
\providecommand{\url}[1]{#1}
\csname url@rmstyle\endcsname
\providecommand{\newblock}{\relax}
\providecommand{\bibinfo}[2]{#2}
\providecommand\BIBentrySTDinterwordspacing{\spaceskip=0pt\relax}
\providecommand\BIBentryALTinterwordstretchfactor{4}
\providecommand\BIBentryALTinterwordspacing{\spaceskip=\fontdimen2\font plus
\BIBentryALTinterwordstretchfactor\fontdimen3\font minus
  \fontdimen4\font\relax}
\providecommand\BIBforeignlanguage[2]{{%
\expandafter\ifx\csname l@#1\endcsname\relax
\typeout{** WARNING: IEEEtran.bst: No hyphenation pattern has been}%
\typeout{** loaded for the language `#1'. Using the pattern for}%
\typeout{** the default language instead.}%
\else
\language=\csname l@#1\endcsname
\fi
#2}}

\bibitem{Kajita:2003uh}
S.~Kajita, F.~Kanehiro, and K.~Kaneko, ``{Biped walking pattern generation by
  using preview control of zero-moment point},'' in \emph{{ICRA}}, 2003.

\bibitem{Herdt:2010bh}
A.~Herdt, N.~Perrin, and P.-B. Wieber, ``{Walking without thinking about it},''
  in \emph{{IROS}}, 2010, pp. 190--195.

\bibitem{Sherikov:2014vq}
A.~Sherikov, D.~Dimitrov, and P.-B. Wieber, ``{Whole body motion controller
  with long-term balance constraints},'' \emph{{IROS}}, 2014.

\bibitem{Feng:2013}
S.~Feng, X.~Xinjilefu, W.~Huang, and C.~Atkeson, ``3d walking based on online
  optimization,'' in \emph{{Humanoids}}, 2013.

\bibitem{Faraji:2014tl}
S.~Faraji, S.~Pouya, and A.~Ijspeert, ``{Robust and Agile 3D Biped Walking With
  Steering Capability Using a Footstep Predictive Approach},'' in
  \emph{{R:SS}}, Berkeley, USA, July 2014.

\bibitem{herzog:2014b}
A.~Herzog, L.~Righetti, F.~Grimminger, P.~Pastor, and S.~Schaal, ``{Balancing
  experiments on a torque-controlled humanoid with hierarchical inverse
  dynamics},'' in \emph{{IROS}}, 2014.

\bibitem{Englsberger:2015jp}
J.~Englsberger, C.~Ott, and A.~Albu-Schaffer, ``{Three-Dimensional Bipedal
  Walking Control Based on Divergent Component of Motion},'' \emph{IEEE
  Transactions on Robotics}, vol.~31, no.~2, pp. 355--368, 2015.

\bibitem{Audren:2014gl}
H.~Audren, J.~Vaillant, A.~Kheddar, A.~Escande, K.~Kaneko, and E.~Yoshida,
  ``{Model preview control in multi-contact motion-application to a humanoid
  robot},'' in \emph{{IROS}}, 2014, pp. 4030--4035.

\bibitem{Wensing:2014bo}
P.~M. Wensing and D.~E. Orin, ``{Development of high-span running long jumps
  for humanoids},'' in \emph{{ICRA}}, 2014, pp. 222--227.

\bibitem{Dai:2014tp}
H.~Dai, A.~Valenzuela, and R.~Tedrake, ``{Whole-body Motion Planning with
  Simple Dynamics and Full Kinematics},'' \emph{{Humanoids}}, 2014.

\bibitem{Lee:2012hb}
S.-H. Lee and A.~Goswami, ``{A momentum-based balance controller for humanoid
  robots on non-level and non-stationary ground},'' \emph{Autonomous Robots},
  vol.~33, pp. 399--414, 2012.

\bibitem{Herzog:2014uv}
\BIBentryALTinterwordspacing
A.~Herzog, N.~Rotella, S.~Mason, F.~Grimminger, S.~Schaal, and L.~Righetti,
  ``{Momentum Control with Hierarchical Inverse Dynamics on a Torque-Controlled
  Humanoid},'' \emph{Autonomous Robots}, (accepted for publication). [Online].
  Available: \url{http://arxiv.org/abs/1305.2042v1}
\BIBentrySTDinterwordspacing

\bibitem{bretl04}
T.~Bretl, S.~Rock, J.~Latombe, B.~Kennedy, and H.~Aghazarian, ``Free-climbing
  with a multi-use robot,'' in \emph{International Symposium on Experimental
  Robotics ISER}, 2004.

\bibitem{tonneau15}
S.~Tonneau, N.~Mansard, C.~Park, D.~Manocha, F.~Multon, and J.~Pettre, ``A
  reachability-based planner for sequences of acyclic contacts in cluttered
  environments,'' in \emph{International Symposium on Robotics Research ISRR},
  2015.

\bibitem{Orin:2008ge}
D.~E. Orin and A.~Goswami, ``{Centroidal Momentum Matrix of a humanoid robot:
  Structure and Properties},'' in \emph{{IROS}}, 2008.

\bibitem{Gill:2002}
P.~E. Gill, W.~Murray, and M.~A. Saunders, ``Snopt: An sqp algorithm for
  large-scale constrained optimization,'' \emph{SIAM Journal on Optimization},
  vol.~12, no.~4, pp. 979--1006, 2002.

\end{thebibliography}
 
\end{document}